\documentclass[11pt]{article}

\usepackage{acl}
\usepackage{amsmath}
\usepackage{underscore}
\usepackage{placeins}

\usepackage{times}
\usepackage{latexsym}

\usepackage[T1]{fontenc}

\usepackage[utf8]{inputenc}

\usepackage{microtype}
\usepackage{multirow}
\usepackage{inconsolata}
\usepackage{booktabs} 

\usepackage{graphicx}

\title{LIME-LLM: Probing Models with Fluent Counterfactuals, \\Not Broken Text}

\author{
  \textbf{George Mihaila} \\
  University of North Texas \\
  \texttt{\small georgemihaila@my.unt.edu} \And
  \textbf{Suleyman Olcay Polat} \\
  University of North Texas \\
  \texttt{\small suleymanolcaypolat@my.unt.edu} \And
  \textbf{Poli Nemkova} \\
  University of North Texas \\
  \texttt{\small poli.nemkova@unt.edu} \AND
  \textbf{Himanshu Sharma} \\
  University of North Texas \\
  \texttt{\small himanshusharma@my.unt.edu} \And
  \textbf{Namratha V. Urs} \\
  University of North Texas \\
  \texttt{\small namrathaurs@my.unt.edu} \And
  \textbf{Mark V. Albert} \\
  University of North Texas \\
  \texttt{\small mark.albert@unt.edu}
}

\begin{document}
\maketitle
\thispagestyle{plain}

\begin{abstract}

Local explanation methods such as LIME \cite{ribeiro2016lime} remain fundamental to trustworthy AI, yet their application to NLP is limited by a reliance on random token masking. These heuristic perturbations frequently generate semantically invalid, out-of-distribution inputs that weaken the fidelity of local surrogate models. While recent generative approaches such as LLiMe \cite{Angiulli2025LLiMeET} attempt to mitigate this by employing Large Language Models for neighborhood generation, they rely on unconstrained paraphrasing that introduces confounding variables, making it difficult to isolate specific feature contributions. We introduce LIME-LLM, a framework that replaces random noise with hypothesis-driven, controlled perturbations. By enforcing a strict "Single Mask–Single Sample" protocol and employing distinct neutral infill and boundary infill strategies, LIME-LLM constructs fluent, on-manifold neighborhoods that rigorously isolate feature effects. We evaluate our method against established baselines (LIME, SHAP, Integrated Gradients) and the generative LLiMe baseline across three diverse benchmarks: CoLA, SST-2, and HateXplain using human-annotated rationales as ground truth. Empirical results demonstrate that LIME-LLM  establishes a new benchmark for black-box NLP explainability, achieving significant improvements in local explanation fidelity compared to both traditional perturbation-based methods and recent generative alternatives.

\end{abstract}

\section{Introduction}
{\footnotesize 
\textit{``The problem is not that machines think like humans, but that humans think machines think like humans.''} \\ 
\hspace*{\fill} --- Herbert A. Simon}\\

As large language models (LLMs) and transformer-based classifiers have become central to NLP, the need for transparent and trustworthy explanations has grown accordingly \citep{doshi2017rigorous, narayanan2018humans}. Among model-agnostic methods, LIME \cite{ribeiro2016lime} remains widely used due to its simplicity and generality. 
\begin{figure}[t!]
    \centering
    \includegraphics[width=0.38\textwidth]{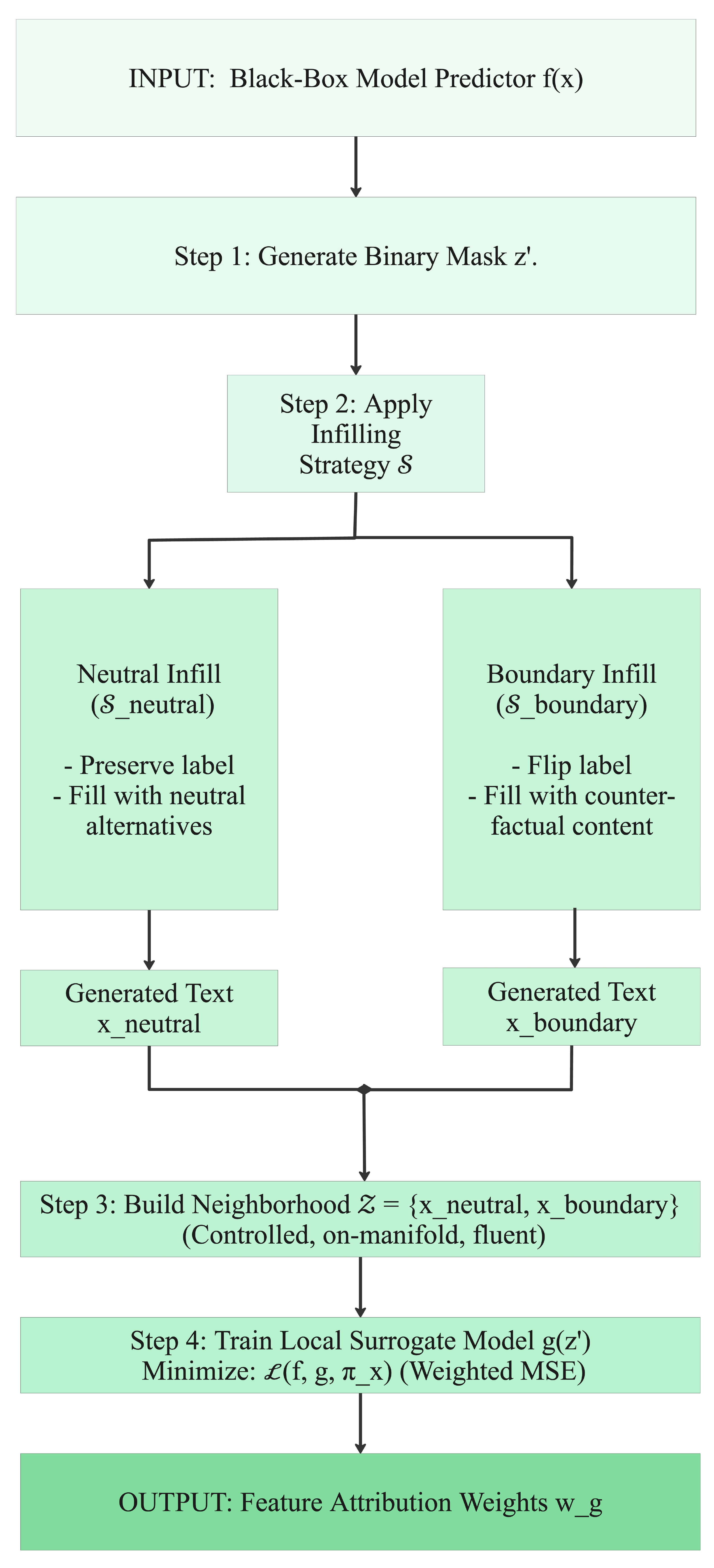}
    \caption{Overview of the proposed LIME-LLM framework.
    LIME-LLM constructs local neighborhoods using hypothesis-driven, on-manifold LLM infilling, generating one fluent sample per binary mask via label-preserving neutral or counterfactual boundary strategies. The resulting semantic neighborhood enables faithful local surrogate explanations. }
    
    \label{fig:flowchart}
\end{figure}
However, in textual settings, LIME constructs local neighborhoods via random token masking, often producing ungrammatical and semantically implausible perturbations that lie far from the data manifold, a limitation repeatedly documented in diagnostic studies \citep{atanasova2020diagnostic, jacovi2020towards}.

Such “broken” inputs lead to unstable explanations and misleading feature attributions \citep{yeh2019fidelity, slack2020fooling}. The issue is particularly acute for tasks where meaning depends on multi-token interactions, including toxicity detection \cite{mathew2021hatexplain}, socially nuanced language \cite{nemkova2023detecting}, and linguistic acceptability, where masking individual tokens often destroys the semantic structure required for reliable local surrogate modeling.

Prior work has sought to improve perturbation-based explanations through phrase-level masking \cite{zhou2025pr} and syntactic constraints \cite{amara2024syntaxshap}. While these approaches improve robustness in specific settings, they typically require additional linguistic resources and remain confined to token-deletion–based perturbation spaces. In parallel, recent work has begun incorporating LLMs into local surrogate frameworks to generate fluent, on-manifold neighborhoods. Notably, LLiMe \cite{angiulli2025llime} employs LLM-based paraphrasing to construct perturbations. However, unconstrained paraphrasing often alters multiple linguistic factors - syntax, lexical choice, and tone - simultaneously, introducing confounding variables that hinder precise feature attribution.

Other directions include game-theoretic attribution \cite{Lundberg2017AUA}, rule-based anchoring \cite{Ribeiro2018AnchorsHM}, and latent-space generation \cite{Lampridis2022ExplainingST}, each with distinct tradeoffs between fidelity, coherence, and computational cost.

In this work, we introduce LIME-LLM, a framework that addresses the tension between semantic validity and feature isolation (Figure \ref{fig:flowchart}). LIME-LLM enforces a strict Single Mask–Single Sample protocol, decoupling hypothesis specification from text generation. For each fixed binary feature mask, the LLM is used strictly as a semantic infiller, generating exactly one controlled perturbation per hypothesis.

We evaluate LIME-LLM against standard model-agnostic baselines, including LIME \cite{ribeiro2016lime} and SHAP , as well as the recent generative baseline LLiMe \cite{angiulli2025llime}. Across global and local evaluation criteria, we show that controlled, hypothesis-driven infilling yields more stable and plausible explanations than either random perturbation or unconstrained generation.

Our results demonstrate that semantic neighborhood construction is a central bottleneck in perturbation-based NLP explainability and that LLM-guided perturbations offer a practical, model-agnostic solution. Empirically, LIME-LLM outperforms standard LIME by 34–135\% in ROC-AUC across tasks and frequently exceeds even white-box Integrated Gradients, despite requiring no access to model internals. By improving explanation faithfulness and stability without sacrificing deployment flexibility, LIME-LLM advances the reliability of local explanations for modern NLP models.


To facilitate reproducibility and community adoption, we release our implementation as an open-source Python package.\footnote{The repository URL will be provided upon acceptance.}

\section{Related Work}
Our work builds on extensive research in post-hoc explainability for NLP, broadly categorized into gradient-based, perturbation-based, and generative model-based methods, which differ in model access assumptions, explanation fidelity, and robustness.

Gradient-based approaches assign token-level importance via model gradients. Methods such as SignedGI \cite{du2021signedgi} and Integrated Gradients \cite{sundararajan2017axiomatic} provide high-fidelity explanations for Transformer classifiers \cite{zheng2024scene} when model internals are accessible, but are inapplicable in black-box or proprietary settings \cite{jacovi2020towards}.

Perturbation-based methods infer importance by modifying inputs and observing prediction changes. LIME \cite{ribeiro2016lime} remains the canonical example, with extensions such as PR-LIME \cite{zhou2025prlime}, XPROB \cite{Cai2024Transparent}, and MExGen \cite{paes2024mexgen} improving phrase coherence, stability, or applicability to generative models. Nonetheless, many still generate semantically implausible neighborhoods, especially for longer or compositional texts.

A parallel line of work addresses the instability of LIME-style explanations, which are sensitive to sampling, kernel width, and surrogate fitting \cite{yeh2019fidelity, slack2020fooling}. Proposed stabilizations include deterministic or clustered neighborhoods DLIME \cite{zafar2019dlime}, hypothesis-testing-based methods S-LIME \cite{zhou2021s}, Bayesian surrogates BayLIME \cite{zhao2021baylime}, and revised sampling or weighting schemes GLIME \cite{tan2023glime}, OptiLIME \cite{visani2020optilime}), all aiming to reduce variance \cite{angiulli2025llime}. However, these methods typically retain token-masking perturbations and thus inherit LIME’s semantic limitations.

Syntax-aware methods explicitly incorporate linguistic structure. SyntaxShap \cite{amara2024syntaxshap} constrains Shapley coalitions using dependency parses, yielding more coherent explanations for autoregressive models such as GPT-2 and Mistral-7B, but typically relies on external parsers and structured representations. Related work also includes SIDU-TXT \cite{Jahromi2024SIDU}, which adapts visual explanation techniques to text, and rule-based grouping approaches such as Anchors \cite{ribeiro2018anchors}.

Generative model-based approaches aim to produce more realistic neighborhoods by sampling from learned latent spaces. Examples include \emph{xspells} \cite{lampridis2022xspells}, which uses variational autoencoders to generate exemplars and counter-exemplars, and XPROA \cite{cai2023xproa}, which applies generative interpolation to approximate local decision boundaries. These methods improve coherence but often require additional training or architectural assumptions.

Most recently, LLM-generated neighborhoods have emerged as a promising alternative. LLiMe \cite{angiulli2025llime} replaces token masking with classifier-driven LLM generation to produce semantically coherent perturbations. While LLiMe mitigates the out-of-distribution problem, it relies on unconstrained paraphrasing that alters multiple linguistic factors simultaneously, introducing confounding variables. We include LLiMe as a generative baseline to demonstrate the advantages of our controlled Single Mask–Single Sample protocol.




\section{Methodology}
We propose LIME-LLM, a model-agnostic explanation framework that addresses the semantic limitations of standard perturbation-based methods. Unlike LIME, which generates local neighborhoods via token deletion, LIME-LLM employs a controlled, hypothesis-driven generation process using LLMs. This approach ensures that all neighborhood samples remain on the data manifold while maintaining the feature isolation necessary for faithful linear approximation.

\subsection{Preliminaries and Problem Formulation}
Let $f: \mathcal{X} \rightarrow [0,1]^C$ be a black-box text classifier that maps an input sequence $x = \{w_1, \dots, w_d\}$ to a probability distribution over $C$ classes. Given a specific instance $x$, LIME seeks to approximate $f$ locally using an interpretable linear model $g(z') = w_g \cdot z'$, where $z' \in \{0,1\}^d$ is a binary vector representing the presence or absence of tokens.

In LIME, a neighborhood $\mathcal{Z}$ is constructed by randomly sampling $N$ binary masks $z'_i$. The corresponding textual representation $z_i$ is created by deleting tokens where $z'_{i,j}=0$. We denote this mapping as $h_{del}(x, z')$. The surrogate $g$ is then trained to minimize the weighted squared loss:
\begin{equation}
\label{eq:lime_loss}
\mathcal{L}(f, g, \pi_x) = \sum_{z'_i \in \mathcal{Z}} \pi_x(z_i) (f(z_i) - g(z'_i))^2
\end{equation}
where $\pi_x(z_i)$ is a proximity kernel (typically exponential cosine distance).

The Manifold Problem: A critical failure mode in NLP is that $h_{del}(x, z')$ frequently yields out-of-distribution (OOD) text (e.g., ``The movie was bad'' to ``movie bad''). Since $f$ was trained on natural text, its predictions on OOD inputs such as $f(z_i)$ are unreliable, introducing noise into Eq. \ref{eq:lime_loss} and degrading the fidelity of $g$.

\subsection{The LIME-LLM Framework}
LIME-LLM replaces the deletion operator $h_{del}$ with a generative operator $G_{LLM}(x, z', \mathcal{S})$, where $\mathcal{S}$ denotes a specific infilling strategy. Our goal is to generate a neighborhood where every sample $x_{new}$ is fluent and on-manifold ($x_{new} \in \mathcal{M}$), yet strictly corresponds to the binary hypothesis $z'$.

The Single Mask - Single Sample Protocol: Recent generative approaches like LLiMe \cite{angiulli2025llime} rely on unconstrained paraphrasing to generate neighborhoods. This introduces feature collinearity: if a paraphrase alters syntax, tone, and vocabulary simultaneously, the linear model cannot isolate the marginal contribution of any single feature. 
To resolve this, we enforce a one-to-one correspondence between the hypothesis space and the text space. For every generated sample $i$:
\begin{enumerate}
    \item We sample a unique binary mask $z'_i$.
    \item We generate exactly one text sample $x_i = G_{LLM}(x, z'_i, \mathcal{S})$.
    \item We enforce that tokens corresponding to $z'_{i,j}=1$ (anchors) are preserved verbatim, while $z'_{i,j}=0$ (masked slots) are infilled.
\end{enumerate}

\subsection{Hypothesis-Driven Infilling Strategies}
To robustly map the decision boundary, we employ two distinct strategies $\mathcal{S}$ that guide the LLM's generation process:

\begin{enumerate}
\item Neutral Infill ($\mathcal{S}_{neutral}$): This strategy probes the robustness of the anchor features. The LLM is instructed to infill masked slots ($z'=0$) with contextually appropriate alternatives that \textit{preserve} the original predicted label $y_{pred} = \text{argmax } f(x)$. 
\begin{equation}
    x_{neutral} \sim P_{LLM}(\cdot | x_{masked}, \text{label}=y_{pred})
\end{equation}
This strategy populates the interior of the local decision cluster, stabilizing the intercept of the surrogate model.

\item Boundary Infill ($\mathcal{S}_{boundary}$): This strategy acts as a decision boundary probe. The LLM is instructed to infill masked slots with alternatives that push the prediction toward a counterfactual label $y_{target} \neq y_{pred}$.
\begin{equation}
    x_{boundary} \sim P_{LLM}(\cdot | x_{masked}, \text{label}=y_{target})
\end{equation}
Significantly, our strategy leverages intelligent masking, which prioritizes the masking of tokens with high prior attribution, such as strong sentiment words, to enable the LLM to effectively reassign labels by providing ample semantic flexibility.
\end{enumerate}

Avoiding the Infilling Trap: A common failure mode in generative XAI is the tendency of LLMs to prioritize fluency over valid label flips (e.g., replacing ``terrible'' with ``poor'' instead of ``great''). We mitigate this via negative constraints in the prompt, explicitly forbidding synonyms for masked high-importance tokens when $\mathcal{S}_{boundary}$ is active.

\subsection{Local Surrogate Optimization}
Once the neighborhood $\mathcal{N} = \{(z'_i, x_i)\}_{i=1}^N$ is generated, we obtain predictions $f(x_i)$ from the black-box classifier. The linear surrogate $g$ is trained by solving:
\begin{equation}
    \hat{w}_g = \operatorname*{argmin}_{w} \sum_{i=1}^N \pi_x(x_i) (f(x_i) - w \cdot z'_i)^2 + \Omega(w)
\end{equation}

\subsection{Hybrid Proximity Kernel}
LIME weights samples using cosine distance on sparse Bag-of-Words (BoW) vectors, which captures lexical overlap but misses semantic proximity. To better capture the decision boundaries of transformer-based models, LIME-LLM employs a hybrid proximity kernel that combines lexical and semantic signals:

\begin{equation}
\pi_{x}(x_{i}) = \frac{1}{2} \left( \text{CosSim}_{bow}(x, x_{i}) + \text{CosSim}_{emb}(x, x_{i}) \right)
\end{equation}
where $\text{CosSim}_{bow}$ measures surface-level token overlap and $\text{CosSim}_{emb}$ measures dense semantic similarity using sentence embeddings. This hybrid weighting ensures that perturbations are penalized not just for changing words, but for drifting from the original semantic meaning. The specific embedding model is detailed in Section \ref{sec:4.4}.



\section{Experiments}
\label{sec:experiments}

To systematically evaluate LIME-LLM, we performed experiments on three diverse text classification benchmarks using BERT-base classifiers \cite{Devlin2019BERTPO}. We compare our approach against perturbation-based, gradient-based, and generative explanation methods.

\subsection{Datasets and Sampling Protocol}
We utilized the following datasets representing three distinct linguistic challenges: sentiment analysis, toxicity detection, and linguistic acceptability.

\begin{enumerate}
    \item \textbf{CoLA} \cite{warstadt2019neural}: A binary acceptability task, with a focus on \textit{Unacceptable} sentences to test the method's ability to pinpoint subtle syntactic violations (e.g., subject-verb disagreement).
    \item \textbf{SST-2} \cite{socher2013recursive}: Binary sentiment classification. Tests the attribution of sentiment-bearing adjectives and modifiers.
    \item \textbf{HateXplain} \cite{mathew2021hatexplain}: A 3-class toxic language task, with a focus on the \textit{Hate} and \textit{Offensive} classes to test sensitivity to context and slurs.

\end{enumerate}

For our final evaluation, we constructed a stratified test set of 150 instances, consisting of 50 samples from each dataset. We prioritize annotation quality over scale, following established practice in human-centered XAI evaluation \cite{DeYoung2019ERASERAB}. To ensure that our explanations address the most critical use cases for interpretability, we applied strict filtering criteria during sample selection:
\begin{itemize}
    \item \textbf{Correctness Constraint:} We selected only instances that were correctly predicted by the black-box classifier. This ensures that our evaluation measures the method's ability to explain the model's valid decision logic rather than its error modes.
    \item \textbf{Utility Constraint (Class Filtering):} We restricted evaluation to classes where the need for explanation is highest. For CoLA, we selected only Unacceptable examples. In detailed linguistic analysis, it is more valuable to explain why a sentence is broken (locating the error) than to explain why a sentence is correct. For SST-2, we utilized balanced Positive/Negative samples. For HateXplain, we excluded the Normal class, focusing exclusively on Hate and Offensive speech. The primary utility of explainable AI in moderation systems is to justify flags for toxicity, not to explain benign content.

\end{itemize}

All test instances are evaluated against human-annotated token-level rationales. For HateXplain, we use the provided ground truth; for SST-2 and CoLA, which lack native token-level labels, we collected human annotations for the test samples (details in Appendix \ref{app:annotation})\footnote{Annotation guidelines provided in the repository, URL will be provided upon acceptance.}.

\textbf{Prompt Tuning \& Development Protocol:} To ensure generalizability, we employed a unified prompt template\footnote{Full prompt can be found in the repository, URL will be provided upon acceptance.} across all tasks, adapting to specific domains solely through a modular \texttt{\{dataset\_description\}} field. We maintained strict train-test separation by optimizing these descriptions on a development set drawn exclusively from the training splits.
Since SST-2 and CoLA lack token-level rationales in their training data, we generated synthetic rationales using Claude Sonnet 4.5 \cite{anthropic_sonnet45_2025}. This generation pipeline was systematically calibrated on HateXplain and refined using a 30-example seed set. We leveraged these development samples to iteratively fine-tune the \texttt{\{dataset\_description\}} instructions - verifying that the LLM correctly distinguished between \texttt{neutral\_infill} (label-preserving) and \texttt{boundary\_infill} (counterfactual) logic, before freezing the prompts for the final evaluation on the human-annotated test set. For the final assessment, the test sets were independently labeled by two human annotators, achieving substantial inter-annotator agreement (Table \ref{tab:dataset_splits}); complete annotation methodology and reliability analyses are detailed in Appendix \ref{app:annotation}.


\begin{table}
  \small
  \begin{tabular}{lll p{3cm}}
    \textbf{Dataset} & \textbf{Train} & \textbf{Test} & \textbf{Annotation} \\
    \hline
    \\
    CoLA       & 50 & 50 & Human (2, $\alpha$=0.64) \\
    SST-2      & 50 & 50 & Human (2, $\alpha$=0.84) \\

    HateXplain & 50 & 50 & Human (3, $\alpha$=0.46) \\
    \hline
  \end{tabular}
  \caption{\label{tab:dataset_splits}
  Dataset splits for development and evaluation.}
\end{table}

\subsection{Baseline Methods}
We compare LIME-LLM against four representative baselines:

\begin{enumerate}
    \item LIME \cite{ribeiro2016lime}: The standard model-agnostic baseline using random token deletion. We use the official implementation with default parameters (kernel width $\sigma=0.75$).
    \item SHAP \cite{lundberg2017unified}: A game-theoretic approach that estimates Shapley values via subset sampling (token deletion).
    \item Integrated Gradients (IG) \cite{sundararajan2017axiomatic}: A white-box gradient attribution method. We include IG as an upper-bound reference for fidelity, assuming access to model internals.
    \item LLiMe \cite{angiulli2025llime}: The primary generative baseline. LLiMe utilizes an LLM to generate neighborhood samples via unconstrained paraphrasing. This comparison highlights the impact of our ``Single Mask - Single Sample'' protocol versus free-form generation.
\end{enumerate}

\subsection{Evaluation Metrics}
We assess explanation quality by measuring how well the derived feature attributions align with human ground-truth rationales.

\textbf{Decision Boundary Mapping (ROC-AUC):} This is our primary metric for global alignment. We treat feature attribution as a ranking problem, computing the Area Under the Receiver Operating Characteristic Curve (ROC-AUC) by comparing the attribution scores against binary human rationales. A higher ROC-AUC indicates that the method correctly assigns higher importance to tokens that humans deem critical to the label. This metric is robust to the scale of the weights and focuses on relative ordering.

\textbf{Precision-Recall (PR-AUC):} We also report the Area Under the Precision-Recall Curve (PR-AUC). Unlike ROC-AUC, PR-AUC is highly sensitive to the sparsity of the explanation. This metric allows us to analyze the \textit{``Manifold Trade-off''} \cite{feng2018pathologies}: while deletion-based methods, such as LIME, often isolate single words (high precision, low recall), on-manifold methods, like LIME-LLM, tend to attribute importance to broader semantic units or phrases (lower precision against sparse human masks, but often higher semantic validity).

\subsection{Implementation Details}
\label{sec:4.4}
For LIME-LLM, we utilize Claude Sonnet 4.5  as the backbone for the sample generation. We generate $N=20$ samples per instance, split evenly between $\mathcal{S}_{neutral}$ and $\mathcal{S}_{boundary}$. While \cite{angiulli2025llime} suggest a ``sweet spot'' of approximately 30 samples for generative samples, we found that our controlled, hypothesis-driven protocol converges more rapidly than free-form paraphrasing. We achieve high performance with just 20 samples, significantly lowering LLM inference costs and latency while minimizing the risk of prompt drift often observed in larger generation batches. For the semantic component of the proximity kernel, we utilize \texttt{all-mpnet-base-v2} \cite{Reimers2019SentenceBERTSE}, selected for its strong performance on semantic similarity benchmarks.
For the LLM ablation study, we additionally evaluate performance using GPT-4.1 \cite{achiam2023gpt} and  Gemini 3 Flash \cite{GoogleGemini2025} to determine if the method's effectiveness is contingent on a specific model architecture discussed in (Additional details in Appendix \ref{app:ablation}).

\begin{figure*}[t!]
    \centering
    \includegraphics[width=0.95\textwidth]{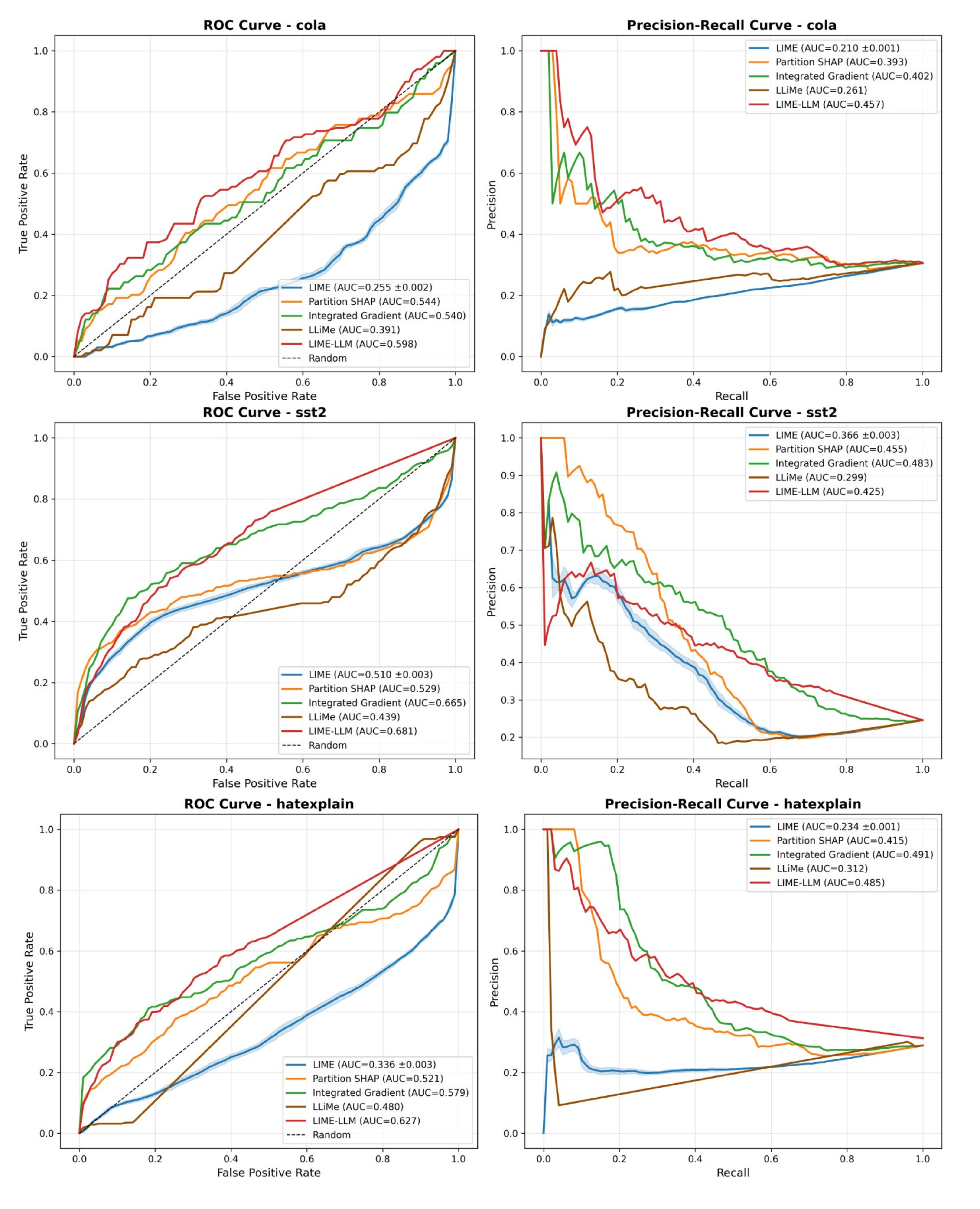}
    \caption{
    Comparison of ROC and Precision-Recall (PR) curves across three evaluation datasets: CoLA (top), SST-2 (middle), and HateXplain (bottom). Shaded regions indicate 95\% confidence intervals over 30 random seeds for stochastic methods. LIME-LLM consistently outperforms LIME and Partition SHAP, and achieves faithfulness comparable to Integrated Gradients while remaining fully model-agnostic.
    }
    
    \label{fig:llmlime-roc-pr}
\end{figure*}

\section{Results and Discussion}

\label{sec:results}
We evaluate the performance of LIME-LLM against LIME, Partition SHAP, Integrated Gradients (IG), and LLiMe. 
Figure \ref{fig:llmlime-roc-pr} and Table \ref{tab:main_results} summarize the Area Under the Curve (AUC) for both ROC and Precision-Recall (PR) metrics across all three datasets.

\begin{table*}
  \centering
  \begin{tabular}{lcccccc}
    \hline
    & \multicolumn{2}{c}{\textbf{SST-2} (Sentiment)} & \multicolumn{2}{c}{\textbf{HateXplain} (Toxicity)} & \multicolumn{2}{c}{\textbf{CoLA} (Syntax)} \\
    \textbf{Method} & \textbf{ROC} $\uparrow$ & \textbf{PR} $\uparrow$ & \textbf{ROC} $\uparrow$ & \textbf{PR} $\uparrow$ & \textbf{ROC} $\uparrow$ & \textbf{PR} $\uparrow$ \\
    \hline
    LIME (Standard) & 0.510 \scriptsize{$\pm 0.003$} & 0.366 \scriptsize{$\pm 0.003$} & 0.336 \scriptsize{$\pm 0.003$} & 0.234 \scriptsize{$\pm 0.001$} & 0.255 \scriptsize{$\pm 0.002$} & 0.210 \scriptsize{$\pm 0.001$} \\
    Partition SHAP & 0.529 & 0.455 & 0.521 & 0.415 & 0.544 & 0.393 \\
    Integrated Gradients & 0.665 & \textbf{0.483} & 0.579 & \textbf{0.491} & 0.540 & 0.402 \\
    LLime (Generative) & 0.439 & 0.299 & 0.480 & 0.312 & 0.391 & 0.261 \\
    LIME-LLM (Ours) & \textbf{0.681} & 0.425 & \textbf{0.627} & 0.485 & \textbf{0.598} & \textbf{0.457} \\
    \hline
  \end{tabular}
  \caption{\label{tab:main_results}
  Comparison of explanation fidelity across datasets. ROC-AUC measures ranking alignment with human rationales; PR-AUC measures sparsity-adjusted precision. Best black-box results in bold. Mean $\pm$ 95\% confidence intervals reported for LIME over 30 random seeds to account for its stochastic nature.
  }
\end{table*}

\subsection{Alignment with Human Rationales (ROC-AUC)}
LIME-LLM consistently achieves the highest ROC-AUC scores across all three benchmarks, outperforming LIME by substantial margins (+34\% on SST-2, +87\% on HateXplain, +135\% on CoLA). 

\paragraph{Comparison with White-Box Methods.}
Notably, LIME-LLM outperforms Integrated Gradients (IG), a method with access to model internals. On SST-2, LIME-LLM achieves an ROC-AUC of \textbf{0.681} compared to IG's 0.665. Similarly, on CoLA, it achieves \textbf{0.598} versus IG's 0.540. This suggests that probing transformer-based classifiers with linguistically valid counterfactuals can provides a better map of the decision boundary than tracing gradients.

Controlled vs. Unconstrained Generation: The comparison with LLiMe highlights the importance of our "Single Mask - Single Sample" protocol. While LLiMe improves over LIME on syntax tasks (CoLA ROC 0.391), it lags significantly behind LIME-LLM on semantic tasks (SST-2 ROC 0.439 vs 0.681). This confirms that unconstrained paraphrasing (LLiMe) introduces confounding variables that obscure the decision boundary, whereas LIME-LLM's controlled infilling strictly isolates feature contributions.

The Manifold Trade-off (Precision vs. Ranking): While LIME-LLM dominates in ranking metrics (ROC-AUC), Precision-Recall (PR-AUC) scores reveal an intrinsic trade-off between semantic validity and sparsity.

As observed in Table \ref{tab:main_results}, white-box methods like IG typically achieve higher PR-AUC (0.483 on SST-2 and 0.491 on HateXplain) than generative methods. This discrepancy arises from the explanation style:
\begin{itemize}
    \item Gradient methods are pinpoint-sparse: They attribute importance to single tokens (e.g., ``bad''), overlapping perfectly with sparse human masks.
    \item Generative methods are semantically diffuse: To maintain fluency, methods like LIME-LLM and LLiMe perturb entire phrases. Consequently, importance is distributed across the phrase, lowering token-level precision.
\end{itemize}
This effect is most severe for LLiMe (PR 0.261 on CoLA), where unconstrained paraphrasing drifts far from the original lexical structure. LIME-LLM (PR 0.457 on CoLA) mitigates this by anchoring the generation, striking a better balance between manifold validity and feature localization.

\subsection{Task-Specific Analysis}

\textbf{SST-2 (Semantics):} LIME-LLM achieves its strongest gains here (ROC 0.681). The gap between LIME-LLM and LLiMe (ROC 0.439) is widest on this task, indicating that sentiment analysis requires precise, localized counterfactuals (e.g., flipping adjectives) rather than whole-sentence paraphrases.

\textbf{HateXplain (Context):} Toxicity detection often depends on specific trigger words. While LIME (ROC 0.336) fails because deleting a slur renders text nonsensical, LIME-LLM (ROC 0.627) successfully replaces slurs with neutral concepts. LLiMe (ROC 0.480) shows improvement over LIME but lags behind LIME-LLM, possibly because unconstrained paraphrasing of toxic content may trigger LLM safety filters or alter contextual cues beyond the intended label manipulation. LLiMe failed to produce valid explanations for 86\% of HateXplain instances (vs. 0\% on SST-2, 4\% on CoLA), likely due to safety filters blocking toxic paraphrases.

\textbf{CoLA (Syntax):} This task validates the "on-manifold" hypothesis. Both generative methods - LIME-LLM (ROC 0.598, PR 0.457) and LLiMe (ROC 0.391, PR 0.261) - drastically outperform Standard LIME (ROC 0.255, PR 0.210). Since CoLA measures linguistic acceptability, "broken" deletion-based samples provide no signal. The success of LLiMe here suggests that for syntax, any fluent perturbation is better than noise, though LIME-LLM's controlled infilling still yields the best fidelity.

\section{Ablation Study}
\label{sec:ablation}

To assess the robustness of LIME-LLM, we perform controlled ablations on two key components: the generative backbone and the locality weighting function. Table \ref{tab:combined_ablation} summarizes the results across all datasets.

\subsection{Backbone Model Sensitivity}
Replacing Claude Sonnet 4.5 with GPT-4.1 \cite{openai2024gpt41} and Gemini 3 Flash \cite{GoogleGemini2025} confirms that our gains are not model-specific. While absolute scores vary, relative improvements remain robust: even when using lower-performing backbones (e.g., GPT-4.1 on semantic tasks), LIME-LLM significantly outperforms Standard LIME on SST-2, while Gemini 3 Flash surpasses the white-box Integrated Gradients baseline on HateXplain (ROC 0.659). This demonstrates that LIME-LLM's efficacy stems from the hypothesis-driven protocol rather than the idiosyncrasies of a single LLM.

\subsection{Distance Metric Sensitivity}
We evaluate three neighborhood weighting kernels: (1) BoW (sparse lexical cosine), (2) Embedding (dense semantic cosine), and 
(3) Hybrid (arithmetic mean). The Hybrid weighting consistently improves fidelity, confirming that the surrogate must penalize both lexical deviation (BoW) and semantic drift (Embedding) to accurately map the transformer's decision boundary. Full tabulation of results for both ablation studies can be found in Appendix \ref{app:ablation}.

\section{Conclusion}

Standard perturbation-based explainers suffer from a fundamental validity problem: probing models with broken, OOD text. We presented LIME-LLM, a framework that addresses this by replacing random deletion with hypothesis-driven semantic infilling. Our key insight is the "Single Mask–Single Sample" protocol, which decouples hypothesis specification from text generation, enabling controlled perturbations that remain fluent while strictly isolating feature contributions.
Through complementary neutral and boundary infill strategies, LIME-LLM constructs on-manifold neighborhoods that probe both the interior and boundaries of local decision regions. Empirically, LIME-LLM outperforms standard LIME by 34–135\% in ROC-AUC across sentiment, toxicity, and syntax tasks, and frequently exceeds white-box Integrated Gradients despite requiring no model internals. Our results suggest that semantic neighborhood construction, not surrogate model complexity, is the central bottleneck in perturbation-based NLP explainability.

\section*{Limitations}
While LIME-LLM substantially improves the semantic validity and stability of local explanations for text, several limitations remain:

\begin{enumerate}
    \item Computational Cost. LLM-guided neighborhood construction incurs higher computational overhead than token-deletion–based methods, which may limit scalability in large-scale or latency-sensitive settings. To mitigate this, LIME-LLM employs a batched prompting strategy in which multiple perturbations are generated within a single LLM invocation. In our experiments, each explanation uses one structured prompt to produce 20 hypothesis-driven perturbations, resulting in 150 total LLM calls for 150 explained instances. While this design substantially improves efficiency compared to per-sample invocation schemes, computational cost remains an open concern. We provide detailed comparison of computation analysis in Appendix \ref{app:cost_calculation}.
    \item Dependence on the Underlying LLM. Explanation quality depends on the behavior of the underlying language model. Although our ablation study shows consistent relative gains across multiple state-of-the-art LLMs, differences in generation style, inductive biases, or safety constraints may affect neighborhood construction, particularly for dialectal, low-resource, or domain-specific language.
      \item Evaluation Scale.Our evaluation is conducted on a relatively small test set (150 instances) compared to large-scale, fully automatic explainability studies. This design choice is intentional. Unlike prior work that evaluates explanation faithfulness using model-based perturbation metrics at scale, our primary objective is to assess semantic alignment with human rationales, which requires expert annotation and careful adjudication. As human rationale annotation is costly and time-intensive, we prioritize depth and annotation quality over breadth. We note that this trade-off is common in human-centered explainability research and that our results should be interpreted as evidence of explanation quality under controlled, high-fidelity evaluation rather than as a large-scale stress test.
    \item Model and Task Coverage. Experiments are currently limited to BERT-style discriminative classifiers on sentence-level tasks. LIME-LLM has not yet been evaluated on non-BERT architectures, generative models, or structured prediction tasks, where perturbation semantics and locality assumptions may differ.
    \item Multilingual and Code-Switched Text. We have not yet systematically examined multilingual or code-switched inputs. Although LIME-LLM is model-agnostic in principle, neighborhood quality may vary across languages depending on LLM coverage and available linguistic resources.
    \item Scalability with Input Length. The scalability of LIME-LLM with respect to input length remains unexplored. Our evaluation focuses on short texts; extending the approach to longer documents or paragraph-level inputs raises open questions regarding perturbation granularity, computational cost, and explanation fidelity.
    \item Locality and Global Faithfulness. As with all local post-hoc explanation methods, LIME-LLM approximates model behavior in the vicinity of individual inputs and does not provide guarantees of global interpretability or causal faithfulness.
  
\end{enumerate}


\section*{Broader Impact and Ethics}
Improving interpretability in NLP systems can support safer and more responsible deployment in domains such as content moderation, decision support, and policy analysis. By generating semantically coherent, on-manifold perturbations, LIME-LLM enables more faithful local explanations, helping practitioners diagnose model behavior, identify spurious correlations, and better understand model responses to sensitive language.

At the same time, post-hoc explanations carry ethical risks. Explanations may be over-trusted or misinterpreted as causal, even though LIME-LLM provides only a local approximation of model behavior. The use of LLMs for neighborhood generation may also introduce stylistic or demographic biases that affect explanation fairness, particularly across dialects or underrepresented language varieties.

More informative explanations can additionally expose sensitive content or be exploited by adversarial actors to probe or evade model behavior, especially in moderation settings. These risks underscore the need to treat explanations as diagnostic tools rather than guarantees of correctness and to pair them with human oversight and broader auditing practices.

Our evaluation involves human-annotated social media data containing toxic or offensive language, which requires careful handling and responsible release. We do not anticipate direct environmental impacts beyond those associated with standard LLM inference used during explanation generation.


\bibliography{custom}

\clearpage
\appendix
\label{sec:appendix}
\onecolumn
\section{Ablation Results}
\label{app:ablation}

\subsection{Ablation Study (Base LLM: GPT-4.1)}
\begin{figure*}[b!]
    \centering
    \includegraphics[width=0.95\textwidth]{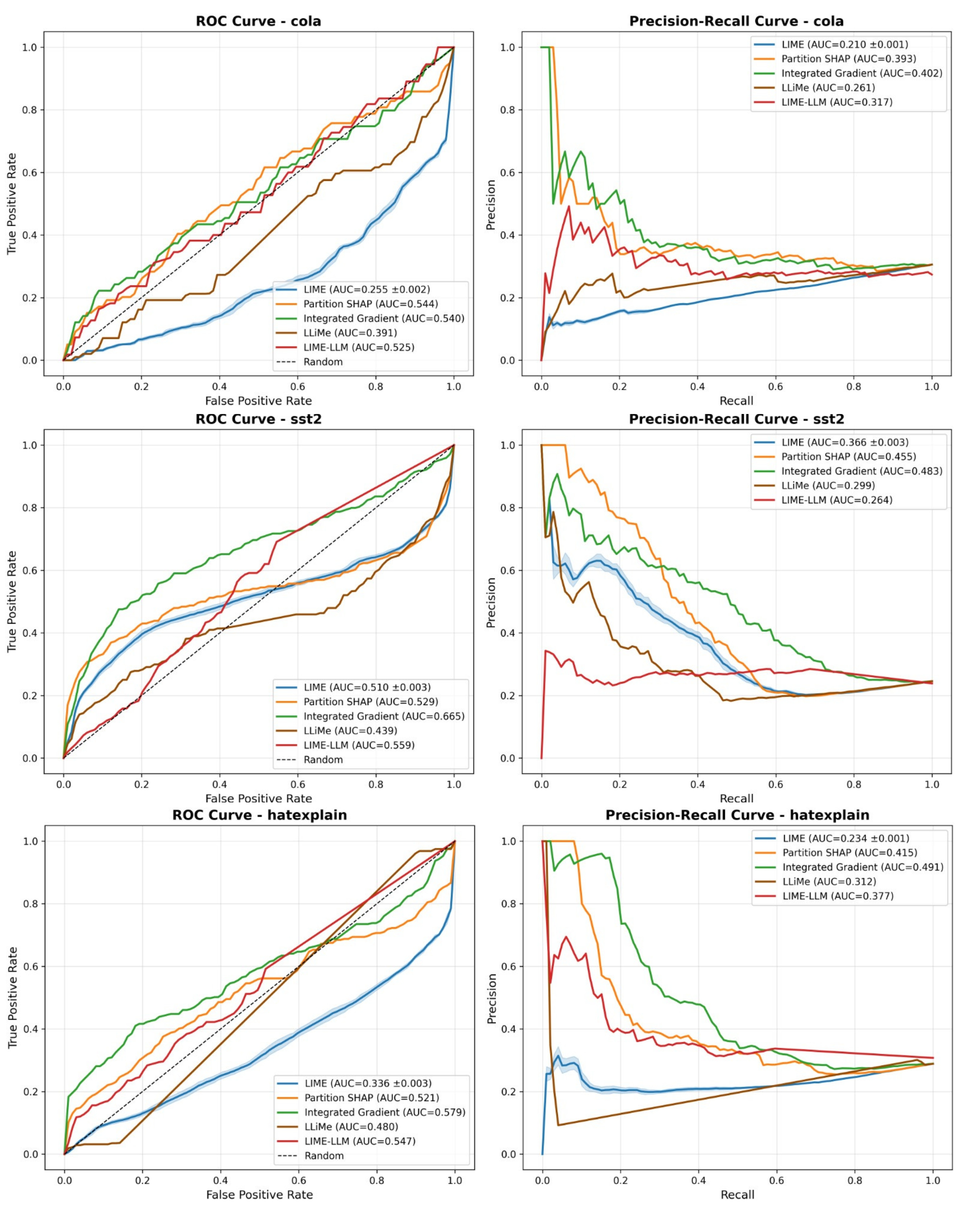}
    \caption{
     Ablation Study 1 (Base LLM: GPT-4.1).
    Comparison of ROC and Precision–Recall (PR) curves across three evaluation datasets: CoLA (top), SST-2 (middle), and HateXplain (bottom). Shaded regions denote 95\% confidence intervals over 30 random seeds for stochastic methods.
    }
    
    \label{fig:ablation1}
\end{figure*}

\newpage
\subsection{Ablation Study (Base LLM: Gemini 3 Flash)}
\begin{figure*}[b!]
    \centering
    \includegraphics[width=0.95\textwidth]{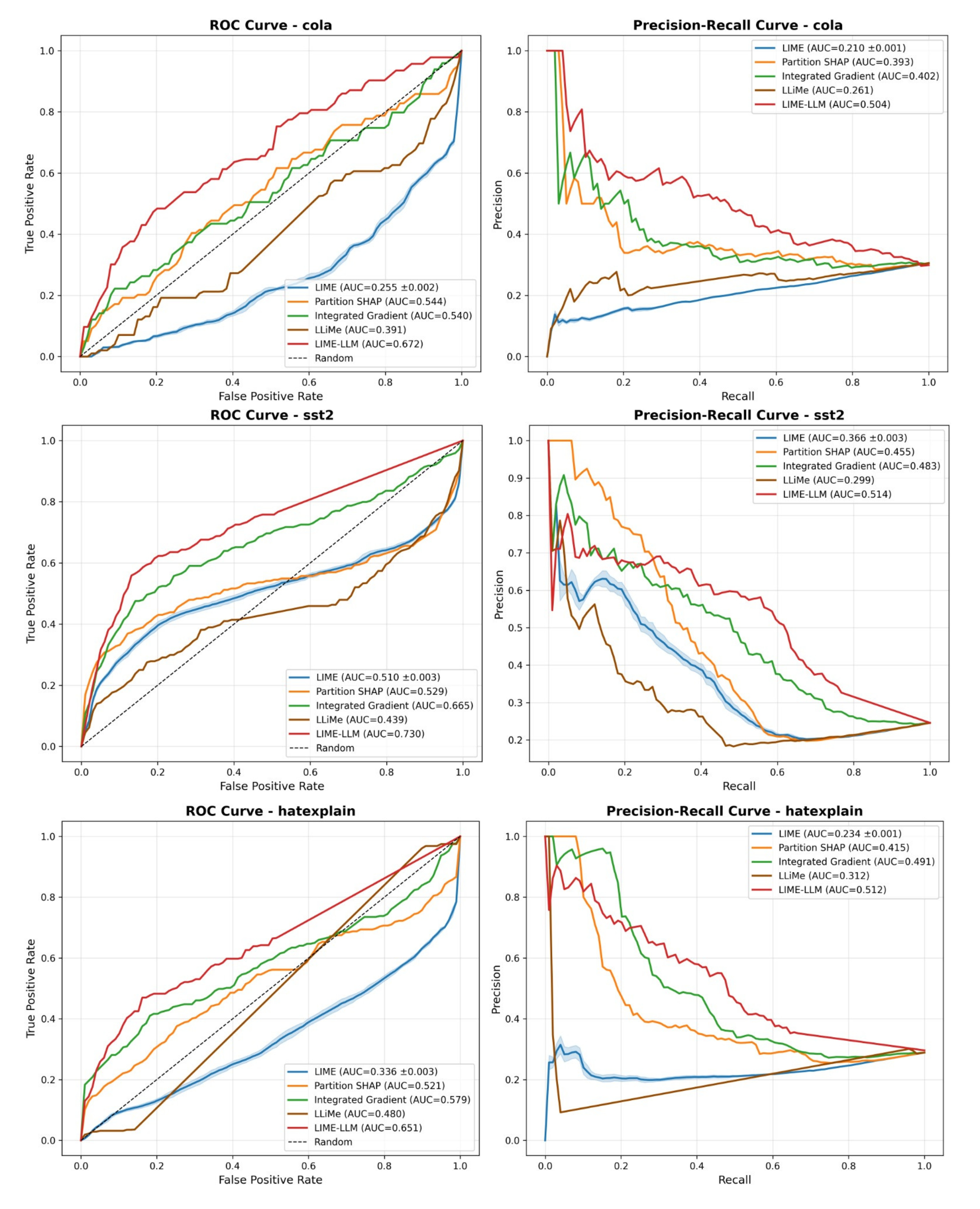}
    \caption{
     Ablation Study 1 (Base LLM: Gemini 3 Flash).
    Comparison of ROC and Precision-Recall (PR) curves across three evaluation datasets: CoLA (top), SST-2 (middle), and HateXplain (bottom). Shaded regions denote 95\% confidence intervals over 30 random seeds for stochastic methods.
    }
    
    \label{fig:ablation2}
\end{figure*}

\newpage
\subsection{Unified Ablation Table}
\begin{table*}[!ht]
\centering
\small
\setlength{\tabcolsep}{4pt} 
\begin{tabular}{l cc c cc}
\toprule
& \multicolumn{3}{c}{\textbf{Backbone Model (ROC-AUC) with Hybrid Distance Metric}} & \multicolumn{2}{c}{\textbf{Distance Kernel (ROC-AUC)}} \\
\cmidrule(lr){2-4} \cmidrule(lr){5-6}
\textbf{Dataset} & \textbf{Claude 4.5 (Main Results)} & \textbf{GPT-4.1} & \textbf{Gemini 3} & \textbf{BoW} & \textbf{Embedding} \\
\midrule
CoLA        & 0.598 & 0.578 & 0.544 & 0.622 & 0.587  \\
SST-2       & 0.681 & 0.578 & 0.682 & 0.655 & 0.697  \\
HateXplain  & 0.626 & 0.550 & 0.659 & 0.611 & 0.599  \\
\bottomrule
\end{tabular}
\caption{Ablation results for LIME-LLM. \textbf{Left:} Sensitivity to the choice of LLM backbone. \textbf{Right:} Comparison of Bag-of-Words, dense embedding, and hybrid average weighting schemes. The hybrid scheme achieves the best performance; all main results (Claude Sonnet 4.5, GPT-4.1 and Gemini 3) are reported using this setting. The method demonstrates high robustness, consistently outperforming Standard LIME baselines regardless of the backbone or weighting scheme.}
\label{tab:combined_ablation}
\end{table*}







\newpage

\section{Annotation Methodology and Reliability}
\label{app:annotation}

\subsection{LLM-Based Annotation Procedure}

For SST-2 and CoLA, which lack ground-truth rationales, we employed a systematic three-stage annotation procedure using large language models:

\paragraph{Stage 1: Calibration on Human-Annotated Data} 
    To ground the procedure in verifiable signal, we calibrated on HateXplain, which provides human token/phrase-level rationales. We evaluated multiple LLMs - \texttt{GPT-4o}, \texttt{GPT-3.5-turbo}, \texttt{Claude-3-5-Haiku-20241022}, and \texttt{Claude-3-5-Sonnet-20241022} - under several prompt variants to predict labels and highlight supporting tokens. We quantitatively compared token-level precision, recall, and F1 against the human spans. Based on this calibration, we identified \texttt{Claude-3-5-Sonnet-20241022} as the most reliable annotator and established best practices for prompt engineering.

\paragraph{Stage 2: Prompt Design and Validation} 
For SST-2 and CoLA, we curated small, manually labeled seed sets with human-annotated rationale examples (30 examples for SST-2, evenly split across classes; 30 examples for CoLA focusing on unacceptable sentences) to anchor the notion of useful token-level rationales. Guided by these examples, we iteratively refined task-specific prompts instructing the LLM to (a) predict the task label, (b) highlight tokens that justify that prediction, and (c) provide a brief explanation for their reasoning. 

We developed seven prompt variants and selected the best-performing version based on the validation sets. For SST-2, the selected prompt achieved micro-F1 = 0.66 and macro-F1 = 0.69 (precision = 0.58, recall = 0.77) on identifying important tokens. For CoLA, we achieved micro-F1 = 0.88 (precision = 0.85, recall = 0.92). These figures reflect expected task differences: sentiment cues are often short and context-dependent (harder to capture consistently), whereas grammatical violations tend to be localized and more systematically recoverable.

\paragraph{Stage 3: Large-Scale Annotation with Quality Controls} 
Using the selected prompts, we annotated the training splits at scale. To avoid propagating rationales for incorrect task predictions, we retained only instances where the LLM's predicted label matched the gold label (label-agreement filter). We then applied light spot-checking to remove outputs with incoherent or diffuse spans. For SST-2, we limited LLM annotation to approximately 6,000 train examples due to API cost constraints. For CoLA, we annotated 2,426 training instances (primarily unacceptable examples). The final annotation counts are: SST-2 (5,971 train / 395 validation), CoLA (2,426 train / 101 validation), and HateXplain (8,042 train / 734 validation), with test sets created by splitting original validation data to ensure label balance.

\begin{table}[ht]
\centering
\footnotesize

\label{tab:llm_calibration}
\begin{tabular}{@{}llccc@{}}
\toprule
\textbf{Dataset} & \textbf{LLM} & \textbf{F1} & \textbf{Seeds} & \textbf{Final} \\
\midrule

\multirow{3}{*}{CoLA}  & claude-3-5-sonnet & \textbf{0.81} & \multirow{3}{*}{30} & \multirow{3}{*}{2,426/101} \\
                        & claude-3-5-haiku  & 0.63 & & \\
                        & gpt-3.5-turbo     & 0.37 & & \\

\midrule
\multirow{3}{*}{SST-2} & claude-3-5-sonnet & \textbf{0.69} & \multirow{3}{*}{30} & \multirow{3}{*}{5,971/395} \\
                        & claude-3-5-haiku  & 0.68 & & \\
                        & gpt-3.5-turbo     & 0.60 & & \\
\midrule
HateXplain & Human & --- & --- & 8,042/734 \\
\bottomrule
\end{tabular}
\caption{LLM annotation calibration and selection results.}
\end{table}

\subsection{Inter-Annotator Agreement Analysis}

To assess the reliability and consistency of our rationale annotations, we computed Krippendorff's Alpha \cite{Krippendorff2011ComputingKA}, a robust inter-annotator agreement coefficient that handles multiple annotators and accounts for chance agreement. For SST-2 and CoLA, we had two human annotators independently verify a subset of LLM-generated rationales on the test sets (195 examples for SST-2, 51 for CoLA). For HateXplain, we report the published inter-annotator agreement from the original dataset \cite{mathew2021hatexplain}, which involved three annotators per example.

\begin{table}[ht]
\centering
\footnotesize
\caption{Inter-annotator agreement and average rationale length (K). For out experiments we limited to 50 samples each to balance the dataset across all 3 datasets and limit the cost of LLM experiment calls.}
\label{tab:interannotator_full}
\begin{tabular}{@{}lccc@{}}
\toprule
\textbf{Dataset} & \textbf{Test Samples} & \textbf{$\alpha$ (Agreement)} & \textbf{Avg. K} \\
\midrule
CoLA        & 51  & 0.64 (moderate)    & 15 \\
SST-2       & 195 & 0.84 (substantial) & 3 \\
HateXplain  & 726 & 0.46 (moderate)    & 9 \\
\bottomrule
\end{tabular}
\end{table}

The higher agreement for SST-2 reflects clearer sentiment-bearing tokens (e.g., ``excellent'', ``terrible''), while CoLA's moderate agreement stems from linguistic nuance in grammatical judgments. HateXplain's lower agreement is consistent with the subjective nature of hate speech annotation, where annotators may focus on different contextual cues. These agreement levels are comparable to or exceed typical values reported in rationale annotation studies \cite{DeYoung2019ERASERAB}, supporting the reliability of our evaluation data.

\newpage
\section{Computational Cost Analysis}
\label{app:cost_calculation}
Table~\ref{tab:computational_cost} reports token consumption and runtime statistics for LIME-LLM and the baseline LLiMe across three LLM backends on the 150-instance test set.

\begin{table*}[!ht]
\centering
\small
\begin{tabular}{lcccc}
\hline
\textbf{LLM Backend} & \textbf{Avg. Input Token} & \textbf{Avg. Output Token} & \textbf{API Calls} & \textbf{Total Runtime} \\
\hline
Claude Sonnet 4.5    & 844 & 1,806 & 150 & 57m 39s \\
GPT-4.1              & 733 & 1,720 & 150 & 52m 47s \\
Gemini 3.0 Flash     & 769 & 748 & 150 & 2h 32m \\
LLiMe                & 155 & 117 & 3,550 & 11h 00m \\
\hline
\end{tabular}
\caption{Token usage and runtime statistics for LIME-LLM across LLM backends. 
Measured on 150-instance test set.}
\label{tab:computational_cost}
\end{table*}


\end{document}